\documentclass{article}

\usepackage{PRIMEarxiv}

\usepackage[utf8]{inputenc} % allow utf-8 input
\usepackage[T1]{fontenc}    % use 8-bit T1 fonts
\usepackage[hidelinks=true]{hyperref}     % hyperlinks
\usepackage{url}            % simple URL typesetting
\usepackage{booktabs}       % professional-quality tables
\usepackage{amsmath,amssymb,amsfonts}    % blackboard math symbols
\usepackage{nicefrac}       % compact symbols for 1/2, etc.
\usepackage{microtype}      % microtypography
\usepackage{fancyhdr}       % header
\usepackage{graphicx}       % graphics

\usepackage{cite}
\usepackage{amsmath,amssymb}
\usepackage{algorithmic}
\usepackage{textcomp}
\usepackage{xcolor}
\makeatletter
\def\bstctlcite{\@ifnextchar[{\@bstctlcite}{\@bstctlcite[@auxout]}}
\def\@bstctlcite[#1]#2{\@bsphack
  \@for\@citeb:=#2\do{%
    \edef\@citeb{\expandafter\@firstofone\@citeb}%
    \if@filesw\immediate\write\csname #1\endcsname{\string\citation{\@citeb}}\fi}%
  \@esphack}
\makeatother

\title{Anatomically Conditioned Recurrent Refinement for Topology-Aware Circle of Willis Segmentation}

\author{
  Juraj Perić$^{1}$\quad Marija Habijan$^{1}$\quad Dario Mužević$^{2}$\quad Irena Galić$^{1}$\quad Danilo Babin$^{3}$\quad Aleksandra Pižurica$^{4}$\\[1.5em]
  $^{1}$Faculty of Electrical Engineering, Computer Science and Information Technology, Osijek, Croatia\\[0.4em]
  $^{2}$Clinical Medical Center Osijek, Osijek, Croatia\\[0.4em]
  $^{3}$Ghent University, Dept.\ of Telecommunications and Information Processing, imec-TELIN-IPI, Ghent, Belgium\\[0.4em]
  $^{4}$Ghent University, Dept.\ of Telecommunications and Information Processing, TELIN-GAIM, Ghent, Belgium\\[0.8em]
  \texttt{juraj.peric@ferit.hr}
}

\begin{document}
\bstctlcite{BSTcontrol}
\maketitle

\begin{abstract}
Segmenting the Circle of Willis (CoW) from Magnetic Resonance Angiography (MRA) is challenging due to complex topology and thin vascular structures that are prone to fragmentation. Standard Convolutional Neural Networks (CNNs) often fail to capture these topological constraints, resulting in "broken vessel" artifacts. To address this, we propose the Anatomically Conditioned Recurrent Refinement U-Net (AC2RUNet). Our architecture decouples segmentation into two streams: a Static Stream that extracts invariant anatomical features and a lightweight Dynamic Stream that iteratively refines topological errors over time. We further introduce a dynamic curriculum learning strategy that transitions from high-recall geometric supervision to topology-aware constraints. Validated on the TopCoW dataset, AC2RUNet substantially reduces Hausdorff Distance (4.72 mm vs 9.17 mm) and Betti number errors (0.19 vs 0.40), improving topological connectivity over the nnU-Net baseline while maintaining comparable volumetric Dice.
\end{abstract}


\section{Introduction}
The Circle of Willis (CoW) is a vital arterial structure located at the base of the brain, responsible for collateral circulation and blood supply to the cerebrum\cite{price_osborns_2014}. Anatomical variations and pathologies within the CoW, such as aneurysms and stenosis, are important indicators for cerebrovascular diseases, including ischemic stroke \cite{liebeskind_collateral_2003,wang_four_2019}. Consequently, accurate segmentation of these vascular structures from Magnetic Resonance Angiography (MRA) is a prerequisite for quantitative analysis, surgical planning, and hemodynamic simulation.

While manual segmentation remains the gold standard, it is labor-intensive and subject to inter-observer variability \cite{yushkevich_user-guided_2006}. In recent years, Convolutional Neural Networks (CNNs), particularly the U-Net \cite{ronneberger_u-net_2015} and its self-configuring variant, nnU-Net \cite{isensee_nnu-net_2021}, have established state-of-the-art performance. Although emerging architectures like Vision Transformers \cite{cao_swin-unet_2021}, Graph Neural Networks \cite{chen_automated_2020,mienye_graph_2025,jia_recent_2025}, and Diffusion Models \cite{wu_medsegdiff_2022,wu_medsegdiff-v2_2023, shi_diffusion_2025} offer powerful global modeling capabilities, they typically require massive datasets or incur prohibitive computational costs that are impractical for scarce MRA annotations. Consequently, the self-configuring nnU-Net remains the most robust baseline for CoW segmentation \cite{isensee_nnu-net_2021, yang_benchmarking_2023}.

However, segmenting the CoW presents unique challenges that standard voxel-wise losses often fail to address. The vascular tree consists of thin, curvilinear structures with high tortuosity and variable contrast. Standard loss functions, such as Cross-Entropy or Dice, optimize for volumetric overlap but lack topological awareness. As a result, predictions often exhibit the "broken vessel" phenomenon, where thin vessels are fragmented, or vital connections are missed. For downstream applications like blood flow modeling, topology is far more critical than pixel-perfect volume; a single disconnection can fundamentally alter the simulated fluid dynamics \cite{shit_cldice_2020}.

Recent approaches have attempted to incorporate topological constraints directly into the loss function. Shit et al. introduced the centerline-Dice (clDice) to penalize disconnected skeletons \cite{shit_cldice_2020}. Similarly, in the recent TopCoW challenge \cite{yang_benchmarking_2023}, several competitors employed Skeleton Recall Loss \cite{kirchhoff_skeleton_2024} to strictly enforce connectivity. While conceptually appealing, recent analyses suggest limitations to this approach, specifically, investigations in \cite{arora_does_2025} have highlighted that these losses can be unstable or fail to generalize across varying vessel radii without careful tuning. Furthermore, standard feed-forward networks often struggle to reconcile these global topological constraints with local intensity details in a single pass. Moreover, incorporating such topological priors often necessitates a trade-off, typically resulting in a degradation of the volumetric Dice score and reduced geometric precision.

To address these limitations, we propose the Anatomically Conditioned Recurrent Refinement U-Net (AC2RUNet). Inspired by \cite{wu_medsegdiff_2022,wu_medsegdiff-v2_2023} and the cognitive workflow of human experts who first localize major structures and then trace fine connections, our model decouples the segmentation task into two streams. A \textit{Static Stream} extracts invariant anatomical features, while a lightweight \textit{Dynamic Stream} iteratively repairs topological errors over time.

In this work we introduce a dual-encoder architecture that decouples heavy feature extraction from iterative refinement, allowing for efficient recurrent inference on 3D volumes. Further, we propose a dynamic curriculum learning strategy that transitions from geometric recall to topological precision, significantly reducing vessel fragmentation.
We demonstrate that AC2RUNet matches the gold-standard nnU-Net on volumetric metrics while substantially improving topological accuracy.

\section{Methodology}
\label{sec:methodology}
\subsection{Network Architecture}
To address topological discontinuities, we formulate CoW segmentation as a recurrent refinement process. Let $X \in \mathbb{R}^{1 \times D \times H \times W}$ denote the input MRA volume. Instead of a single-pass mapping, our proposed AC2RUNet generates a sequence of estimates $\hat{Y}_t$ via the recurrence relation:
\begin{equation}
    \hat{Y}_t = \mathcal{F}(X, \hat{Y}_{t-1}, \theta)
    \label{eq:recurrence}
\end{equation}
where $\mathcal{F}$ is the refinement function parametrized by $\theta$, and the process is initialized with a zero-filled tensor $\hat{Y}_0$. The final segmentation is given by $\hat{Y}_T$.

\begin{figure}[htbp]
\centerline{\includegraphics[width=.8\textwidth]{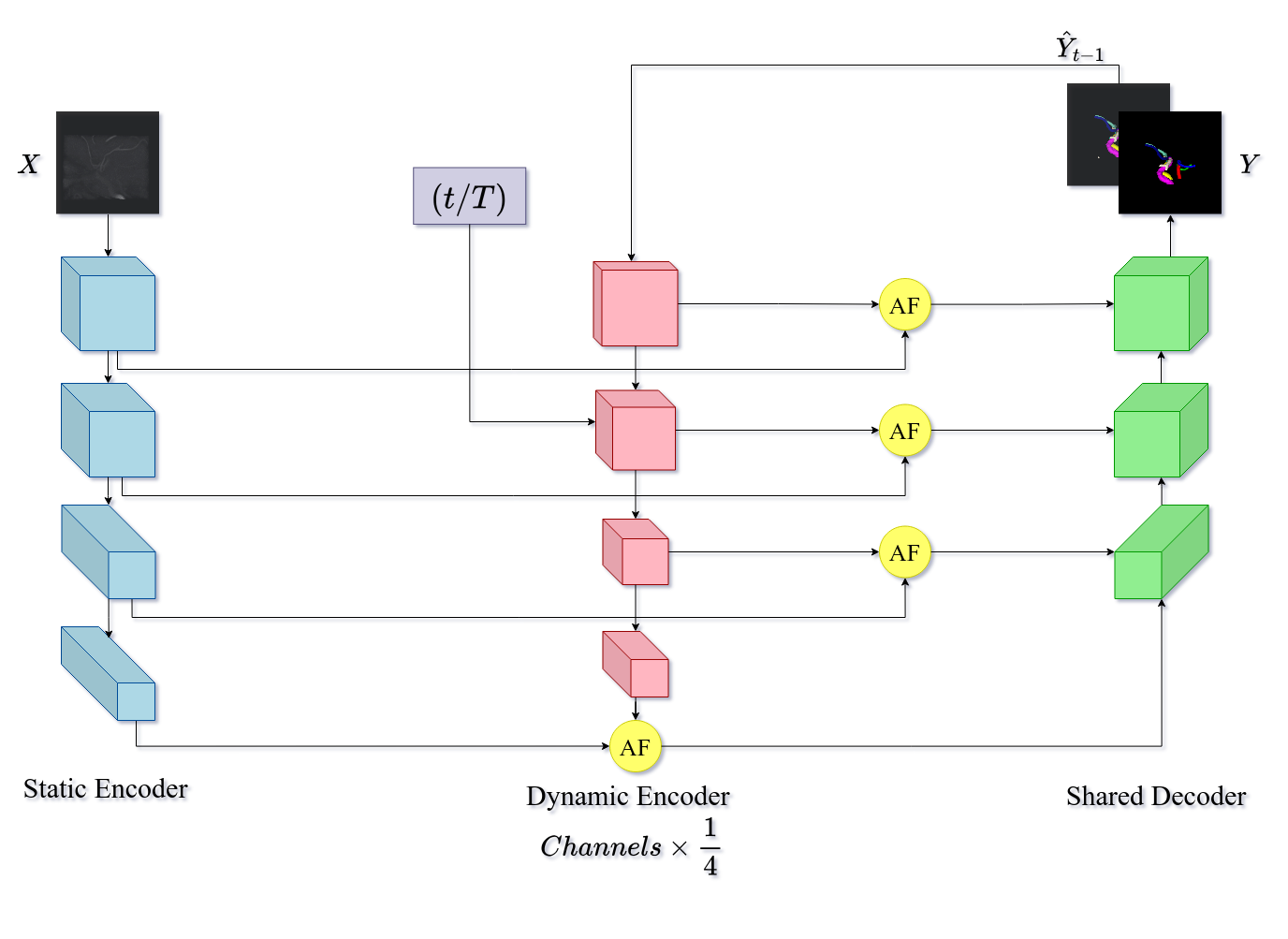}}
\caption{Schematic of AC2RUNet. The Static Encoder (blue) extracts invariant features from input $X$, while the lightweight Dynamic Encoder (pink) refines the previous estimate $\hat{Y}_{t-1}$. Temporal embeddings ($t/T$) and Asymmetric Fusion (AF) modules integrate these signals before the Shared Decoder (green) generates the next estimate, repeating for $T$ steps.}
    \label{fig:architecture}
\end{figure}

To efficiently model this recurrence, we introduce a Dual-Encoder design that decouples processing into two pathways. A \textit{Static Stream} extracts invariant anatomical features from $X$ (computed once at $t=0$), while a lightweight \textit{Dynamic Stream} iteratively processes the evolving probability map $\hat{Y}_{t-1}$. 
Although plain convolutions often outperform residual baselines on small datasets\cite{isensee_nnu-net_2024}, we explicitly adopt a residual design (ResEnc M) to facilitate gradient propagation through temporal sequences, effectively mitigating the vanishing gradient problem inherent in recurrent training\cite{he_deep_2015, alom_recurrent_2018}. Figure \ref{fig:architecture} represents further described architecture.

\subsubsection{Static Anatomical Stream}
This encoder is responsible for extracting invariant structural features from the raw MRA volume $X$. As previously noted, this component is architecturally derived from the encoder pathway of the nnU-Net ResEnc M \cite{isensee_nnu-net_2024}. Structurally, the encoder is composed of a series of stacked Residual Blocks. Each block consists of two $3\times 3 \times 3$ convolutional layers, each followed by Instance Normalization and a Leaky ReLU activation function. A shortcut connection aggregates the input of the block with the output of the second normalization layer before the final activation, facilitating the preservation of feature identity.

Operationally, this encoder is designed for efficiency. It processes the input volume only once at the initialization step $(t=0)$. This single forward pass generates a hierarchical set of multi-scale feature maps, denoted as $h^{(l)}_{stat}$ where $l$ represents the resolution level. These feature maps are cached in memory and serve as a fixed, invariant anatomical basis for all subsequent refinement iterations $(t>0)$, thereby decoupling heavy feature extraction from the iterative refinement loop.

\subsubsection{Dynamic Refinement Stream}
Parallel to the static stream, the Dynamic Stream encoder is tasked with processing the evolving segmentation state. Since this module is executed iteratively ($T$ times) for every single inference, computational efficiency is paramount. To mitigate the computational burden, we designed this encoder to be lightweight; specifically, its channel capacity is reduced by a factor of $\gamma=4$ relative to the static encoder (i.e., base width of 8 vs. 32).

At any given time step $t$, the input to this stream is the probabilistic estimate from the previous iteration, denoted as $P_{t-1}=Softmax(\hat{Y}_{t-1})$. By operating on probability maps rather than raw logits, we ensure the input is bounded in the range $[0,1]$, providing a stable signal for the encoder.

To enable the network to distinguish between early coarse predictions and late fine-tuning stages, we introduce a Time Embedding mechanism. The current step $t$ is normalized by the total steps $T$ (i.e., $\tau=t/T$), ensuring the signal remains agnostic to the specific number of iterations chosen during inference. This scalar $\tau$ is projected into the channel dimension via a small Multi-Layer Perceptron (MLP) and injected into the feature stream via element-wise addition. It is important to say that this injection occurs after the first residual block of the dynamic encoder. This architectural choice prevents the temporal signal from corrupting the raw spatial geometry of the input probability map, allowing the first block to extract pure spatial features before temporal conditioning is applied.

\subsubsection{Asymmetric Feature Fusion}
To integrate the invariant anatomical information with the evolving segmentation state, we employ a simple Asymmetric Feature Fusion mechanism at each resolution level. Since the dynamic stream operates with a reduced channel capacity ($\gamma=4$), a direct element-wise summation is dimensionally impossible. Therefore, we project the dynamic features into the static feature space before fusion.

Formally, let $h^{(l)}_{stat} \in \mathbb{R}^{C\times D \times H\times W}$ denote the cached features from the static encoder, and $h^{(l)}_{dyn, t} \in \mathbb{R}^{\frac{C}{4}\times D \times H\times W}$ represent the features from the dynamic encoder at time step $t$. The fused feature map $Z^{(l)}_t$ is computed as:
\begin{equation}
    Z^{(l)}_t = h^{(l)}_{stat} + \mathcal{F}_{\phi}(h^{(l)}_{dyn, t})
    \label{eq:asymetric_feature_fusion}
\end{equation}
where $\mathcal{F}_{\phi}$ represents a projection function consisting of a $1\times 1\times 1$ convolution followed by Instance Normalization and a Leaky ReLU activation.

Functionally, this fusion acts as a grounding mechanism, suppressing hallucinations by forcing the decoder to reconcile dynamic topological suggestions with the raw anatomical intensity preserved in the static stream.
\subsubsection{Decoder}
The decoder pathway mirrors the geometry of the static encoder, progressively recovering spatial resolution through trilinear upsampling followed by convolutional refinement. At each resolution level, the upsampled features are concatenated with the corresponding fused feature maps $Z^{(l)}_t$ from the fusion module.

To further facilitate gradient flow and enforce structural consistency at multiple scales, we employ Deep Supervision \cite{isensee_nnu-net_2021, isensee_nnu-net_2024} by computing auxiliary losses at the three highest resolution levels, as detailed in the loss function formulation in the next section.

\subsection{Iterative Inference Process}
Inference proceeds in two phases to maximize stability and efficiency:
\begin{enumerate}
    \item \textbf{ Initialization ($t=0$):} the Dynamic Stream is inactive and the network functions as a standard U-Net, where the cached static features $h_{stat}$ pass directly to the decoder to produce an initial coarse segmentation map $\hat{Y}_0$, establishing the general vascular geometry.
    \item \textbf{ Recurrent Refinement ($t>0$):} for $t=1 \dots T$, the Dynamic Stream activates and takes the softmax probabilities of the previous estimate $\hat{Y}_{t-1}$ as input, iteratively refining the segmentation by specifically targeting gaps inside the vessel segmentation.
\end{enumerate}

\subsection{Dynamic Curriculum Learning}

\label{sec:curriculum}

Training a recurrent network to refine thin vascular structures is notoriously unstable if the model is forced to learn complex topology before it understands basic geometry. To mitigate this, we employ a multi-dimensional curriculum that evolves over three axes: recurrent inference steps ($t$), spatial resolution scales ($s$), and training epochs.

\subsubsection{Progressive Spatiotemporal Loss}
We formulate the objective function as a weighted sum over time and scale, prioritizing the final high-resolution output while maintaining gradient flow through intermediate states. The total loss is defined as:

\begin{equation}
\mathcal{L}_{total} = \sum_{t=0}^{T} \omega_{temp}^{(t)} \cdot \mathcal{L}_{hybrid}(\hat{Y}_t, Y)
\end{equation}

where: \begin{itemize}
\item $\omega_{temp}^{(t)}$ are Temporal Weights that increase exponentially $(\propto 2^t)$ to prioritize the final refined prediction.
\item $\mathcal{L}_{hybrid}$ aggregates the geometric and topological loss functions across valid spatial resolutions (via Deep Supervision)
\end{itemize}

\subsubsection{Loss Composition Strategies} 
We investigate two distinct formulations for the component $\mathcal{L}_{hybrid}$ to determine the optimal signal for topological restoration. In both configurations, we employ Deep Supervision \cite{isensee_nnu-net_2021} to enforce geometric consistency across the three highest resolution levels. Following the standard scheme, the loss weights decrease by a factor of 2 at each lower resolution scale (i.e., $\omega_{spat} \propto 2^{-s}$), ensuring that the primary gradient signal is driven by the full-resolution output. However, to maintain computational stability, the topological component is strictly limited to this full-resolution output, meaning the auxiliary losses at lower resolutions consist solely of the geometric terms.

\paragraph{Strategy A: Progressive Dice-Topology}
This approach utilizes standard Soft Dice and Cross-Entropy (CE) as the geometric foundation. To enforce connectivity, we introduce the Soft Centerline Dice (clDice) loss via a progressive schedule $\lambda_{cl}(t)$:

\begin{equation}
\mathcal{L}^{(t)}_A = \mathcal{L}_{Dice} + \mathcal{L}_{CE} + \lambda_{cl}(t)\mathcal{L}_{clDice}
\end{equation}

The schedule is defined as: 
\begin{itemize} 
\item \textbf{Step $t=0$ (Geometry Focus):} $\lambda_{cl}=0$. The network focuses solely on volumetric overlap to establish the vessel bulk.
\item \textbf{Step $t=1$ (Transitional):} $\lambda_{cl}=0.2$. A mild topological penalty is introduced. 
\item \textbf{Step $t=2$ (Topology Focus):} $\lambda_{cl}=0.5$. The penalty is maximized.
\end{itemize}

\paragraph{Strategy B: Recall-Oriented Tversky Curriculum} 
This strategy dynamically manipulates the precision-recall trade-off. We hypothesize that thin vessels are easily missed as False Negatives in the initial static prediction. We mitigate this using the Tversky Index (TI)\cite{salehi_tversky_2017} with evolving False Positive penalty and False Negative penalty parameters as our geometric foundation:

\begin{equation}
\mathcal{L}^{(t)}_{B} = (1 - TI^{(t)}) + \mathcal{L}_{CE} + \lambda_{cl}(t)\mathcal{L}_{clDice} 
\end{equation}

\begin{itemize} 
\item \textbf{Step $t=0$ (High Recall):} $\alpha=0.3, \beta=0.7, \lambda_{cl}=0$. By penalizing False Negatives heavily, the Static Stream is forced to capture faint vessels, ensuring no anatomical components are missed even at the cost of over-segmentation. 
\item \textbf{Step $t=1$ (Balancing):} $\alpha=0.4, \beta=0.6, \lambda_{cl}=0.2$. The network begins to prune False Positives while introducing topological constraints. 
\item \textbf{Step $t=2$ (High Precision):} $\alpha=0.5, \beta=0.5, \lambda_{cl}=0.5$. The final refinement focuses on precise boundary delineation (Standard Dice) and connectivity. 
\end{itemize}
\subsubsection{Staged Training Schedule}
To ensure stability, we employ a three-stage training regimen that gradually increases task complexity. Initially, \textbf{Static Pre-training} (0--20\% epochs) activates only the Static Stream ($t=0$) to allow the heavy encoder to converge on anatomical features. Subsequently, a \textbf{Warm-up Refinement} phase (20--40\% epochs) introduces a single refinement step ($T=1$) to learn initial error correction. Finally, the model transitions to \textbf{Full Recurrence} (40--100\% epochs) with the complete recurrent loop ($T=2$), enabling the learning of complex, iterative topological repairs.

\subsection{Implementation details}
We used 125 MRA images from TopCoW challenge dataset, 100 for training and 25 for testing. Following the standard protocol of the TopCoW Challenge\cite{yang_benchmarking_2023}, we trained our models specifically on the Region of Interest (ROI) crops of the CoW rather than the full field-of-view, as the target vessels occupy a minute fraction of the total cerebral volume. We utilized the self-configuring nnU-Net framework \cite{isensee_nnu-net_2021} to automatically determine the dataset fingerprint. Based on this analysis, all MRA volumes were resampled to a common voxel spacing and normalized via z-score normalization.

The network was trained on 3D patches of size $64\times 192 \times 160$ voxels. To accommodate the recurrent architecture's memory footprint, we used gradient accumulation (micro-batch size 1, 2 steps) to match the standard nnU-Net batch size of 2, doubling the epoch length to 500 iterations to preserve the effective number of parameter updates.
We adhered to the default nnU-Net training scheme\cite{isensee_nnu-net_2021} for 1000 epochs, using SGD with Nesterov momentum ($\mu=0.99$, weight decay $3\times 10^{-5}$) and a poly learning rate schedule ($\eta_0=0.01$). All experiments were conducted on a single NVIDIA RTX 5080 GPU.

\section{Results and Discussion}
We note that evaluation is conducted on the single-center TopCoW MRA cohort (n=125); reported standard deviations should be interpreted with this scale in mind.

\begin{table*}[htbp]
\centering
\caption{Model Performance Metrics}
\label{tab:results}
\resizebox{\textwidth}{!}{
    \begin{tabular}{llccccccc}
    \toprule
    \textbf{Method} & \textbf{Loss} & \textbf{T} & \textbf{Dice (\%)} $\uparrow$ & \textbf{HD95 (mm)} $\downarrow$ & \textbf{clDice (\%)} $\uparrow$ & \textbf{$\beta_0$ Error} $\downarrow$ & \textbf{F1 Grp2 (\%)} $\uparrow$\\
    \midrule
    nnUNet & DiceCE & - & $83.07 \pm 18.09$ & $9.1756 \pm 8.7130$ & $93.14 \pm 4.34$ & $0.4048 \pm 0.9442$ & $77.36 \pm 27.34$ \\
    nnUNet & DiceCE + clDice & - & $82.46 \pm 19.14$ & $8.2215 \pm 9.7769$ & $93.36 \pm 3.61$ & $0.3895 \pm 1.1694$ & $81.55 \pm 24.45$ \\
    \midrule
    AC2RUNet & DiceCE & 3 & $82.24 \pm 17.43$ & $9.3797 \pm 8.2016$ & $92.29 \pm 4.54$ & $0.3875 \pm 0.9237$ & $75.00 \pm 27.49$ \\
    AC2RUNet & DiceCE + clDice & 3 & $83.08 \pm 17.58$ & $8.2266 \pm 9.7412$ & $93.08 \pm 5.02$ & $0.2526 \pm 0.6375$ & $78.00 \pm 35.74$ \\
    AC2RUNet & Tversky + clDice & 3 & \textbf{83.16} $\pm$ 15.76 & \textbf{4.7232} $\pm$ 5.7783 & \textbf{93.39} $\pm$ 4.73 & \textbf{0.1855} $\pm$ 0.7123 & \textbf{84.09} $\pm$ 29.90 \\
    \midrule
    \midrule
    AC2RUNet & Tversky + clDice & 2 & $83.08 \pm 15.74$ & $4.7322 \pm 5.8183$ & $93.38 \pm 4.73$ & $0.2000 \pm 0.7675$ & $84.09 \pm 29.11$ \\
    AC2RUNet & Tversky + clDice & 5 & $83.26 \pm 15.71$ & $4.7267 \pm$ 5.7692 & \textbf{93.56} $\pm$ 4.02 & $0.2145 \pm 0.7683$ & $85.39 \pm 25.01$ \\
    AC2RUNet & Tversky + clDice & 7 & \textbf{83.33} $\pm$ 15.69 & $4.7342 \pm 5.7689$ & $93.54 \pm 4.00$ & $0.1964 \pm 0.7708$ & \textbf{86.67} $\pm$ 24.92 \\
    AC2RUNet & Tversky + clDice & 10 & \textbf{83.33} $\pm$ 15.67 & $4.7433 \pm 5.7735$ & $93.51 \pm 4.05$ & $0.2000 \pm 0.8485$ & \textbf{86.67} $\pm$ 24.92 \\
    
    \bottomrule
    
    \end{tabular}
}
\end{table*}
\subsection{Quantitative Results and Ablation Study}

As presented in Table \ref{tab:results}, the proposed AC2RUNet demonstrates comparable geometric accuracy to the strong nnU-Net baseline while offering distinct improvements in topological continuity. While the standard nnU-Net achieves a Dice score of 83.07\%, our AC2RUNet (configured with Tversky + clDice loss at $T=3$) maintains this high standard at 83.16\% supported by reduction in standard deviation. More importantly, we successfully mitigate the common trade-off between volumetric accuracy and topological consistency. Where traditional topology-aware losses often degrade Dice scores (cf. nnU-Net+clDice dropping to 82.46\%), our method preserves volumetric Dice while improving clDice to 93.39\% over the baseline's 93.14\%.

Table \ref{tab:results} also details the impact of the iterative refinement steps ($T$) on model performance. While $T=3$ was selected as the optimal operating point to balance accuracy and inference speed, the architecture demonstrates remarkable stability across varying timesteps. Increasing $T$ from 2 to 10 yields marginal but consistent gains, with Dice scores peaking at 83.33\% and topological metrics stabilizing further. This is also accompanied by a slight reduction in standard deviation across most metrics, notably in HD95 and $\beta_0$ Error, suggesting that extended refinement further stabilizes the topological reconstruction of small vessels. This trend is consistent across all investigated loss combinations.

\subsection{Topological Robustness and Small Vessel Accuracy}

The most significant improvements are found in the distance and topology-based metrics. The HD95 is reduced by nearly 50\% compared to the baseline (9.17 mm vs. 4.72 mm). This improvement is particularly important for CoW, where vessel diameters  range from 1 mm to 4.6 mm, meaning the proposed model brings the boundary error within the scale of the anatomical structures themselves. Furthermore, the $\beta_0$ Error, which measures the number of disconnected components in the vascular tree, reaches its lowest value of 0.1855 with our curriculum. Two other methods have also shown improvement in $\beta_0$ Error.

These metrics, combined with the reduction in standard deviation for HD95 and Betti error, indicate that the model is far more consistent in its segmentation of small, peripheral vessels. This is further supported by the F1 Grp2 score, which specifically targets smaller vascular structures. Our method achieves 84.09\%, a significant jump from the 77.36\% achieved by the baseline.
\begin{figure}[htbp]
\centerline{\includegraphics[width=.8\textwidth]{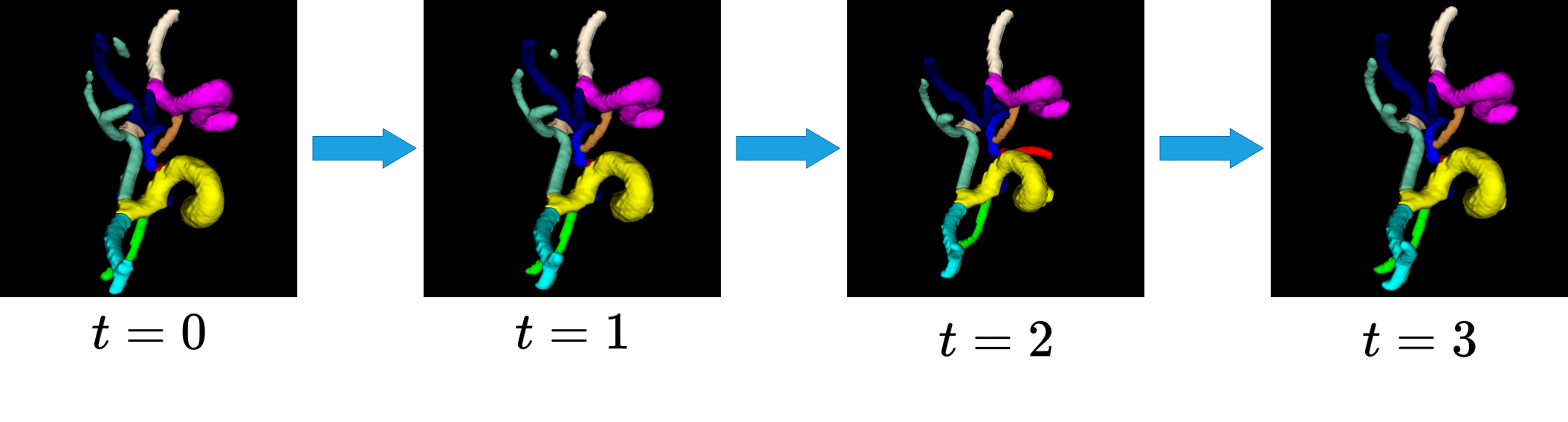}}
\caption{Iterative Refinement Process}
    \label{fig:progress}
\end{figure} 

\subsection{Qualitative Analysis}
Qualitative results confirm AC2RUNet's ability to repair topological defects. Figure \ref{fig:progress} illustrates the iterative recovery of vessel connectivity: initial gaps in thin branches at t=0 are progressively closed by the dynamic stream by t=3, restoring continuity without over-segmentation.

\begin{figure}[htbp]
\centerline{\includegraphics[width=.9\textwidth]{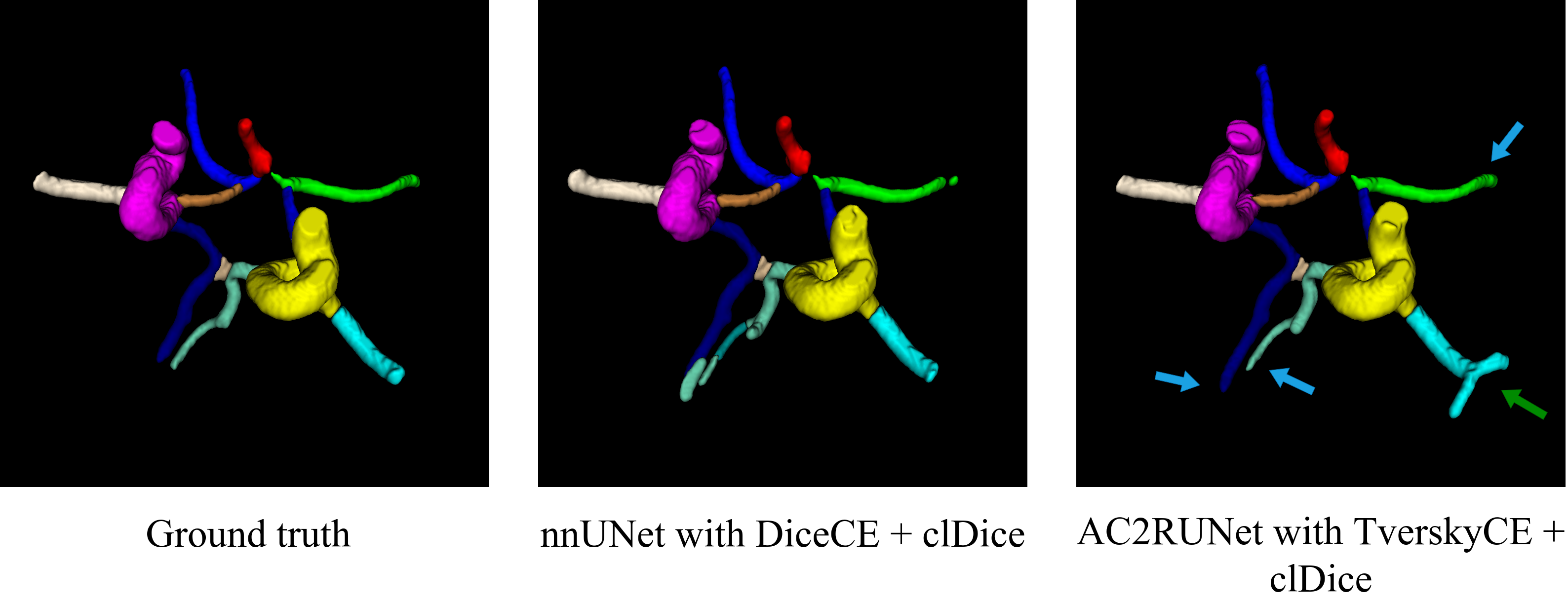}}
\caption{Method comparison}
    \label{fig:method_comparison}
\end{figure}

\begin{figure}[htbp]
\centerline{\includegraphics[width=.75\textwidth]{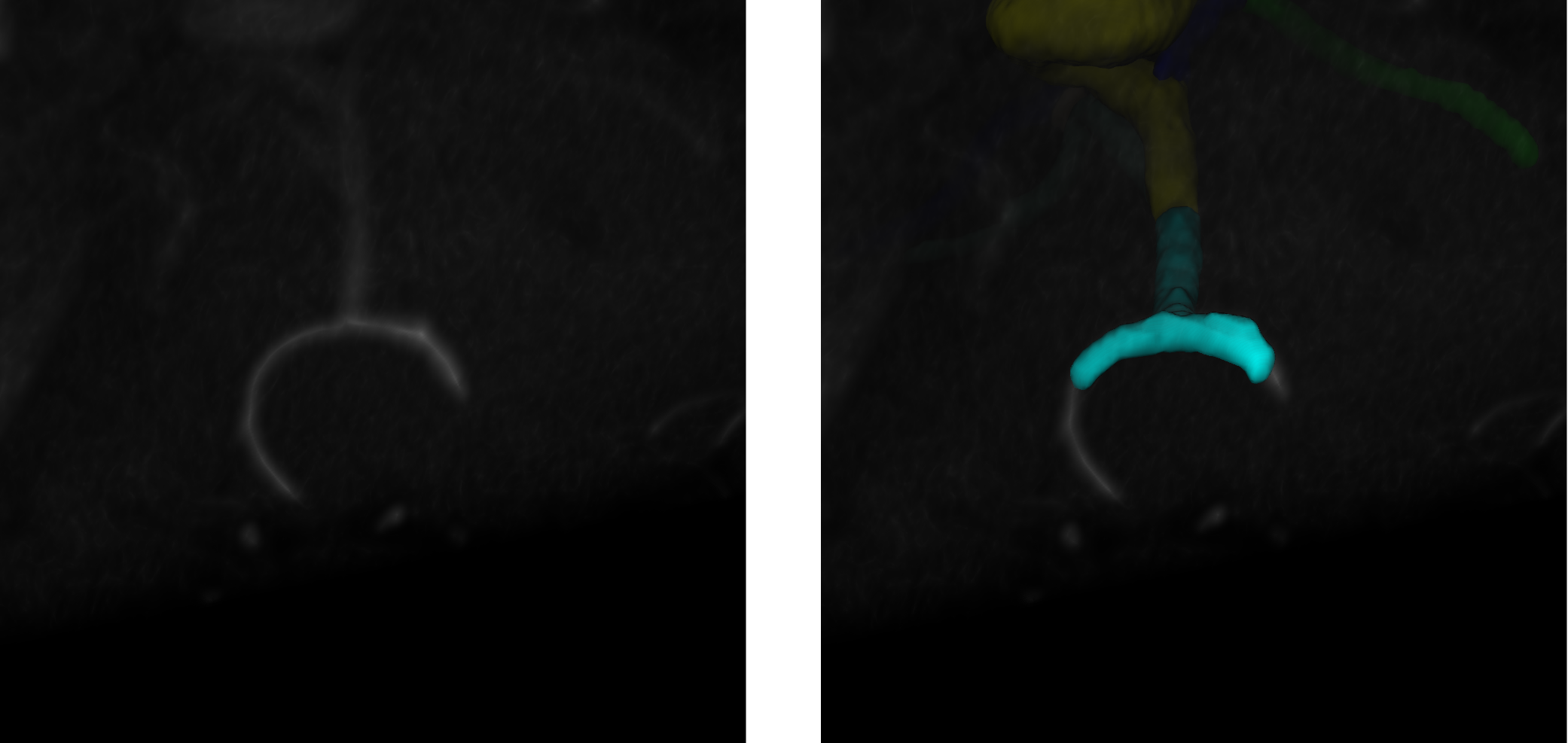}}
\caption{Raw image and AC2RUNet's added segment}
    \label{fig:missing_gt}
\end{figure}

Comparison with the nnU-Net baseline (Figure \ref{fig:method_comparison}) highlights the precision of our Strategy B (Tversky + clDice). Blue arrows indicate where AC2RUNet recovers distal branches and connections missed by the baseline. Notably, the model also produces a segmented branch absent from the ground truth (green arrow). The raw MRA intensity in Figure \ref{fig:missing_gt} shows vessel-like signal at this location, suggesting either a missed annotation or a plausible topology-driven hallucination; we present this case as illustrative rather than evaluative, as definitive interpretation requires expert radiological review.

\subsection{Computational and Memory Efficiency}
AC2RUNet maintains efficiency via a lightweight Dynamic Encoder (5.65M parameters). Each additional iteration requires only 385.68 GFLOPs, yet the memory footprint remains constant regardless of the number of steps ($T$). This decoupling enables superior topological segmentation on VRAM-constrained hardware by exchanging a small increase in computation time for higher accuracy. 

\section{Conclusion and Discussion}
In this work, we introduced the Anatomically Conditioned Recurrent Refinement U-Net (AC2RUNet) to address the unique topological challenges of segmenting the Circle of Willis. By decoupling heavy feature extraction from iterative refinement, we successfully mitigated the "broken vessel" phenomenon, aligning our approach with the cognitive workflow of human experts who prioritize structural localization before detailed tracing.

Our experiments demonstrate that this dual-stream architecture, combined with a dynamic curriculum of Tversky and clDice losses, substantially reduces the Hausdorff Distance (4.72mm from 9.17mm) and Betti number errors compared to the nnU-Net baseline. This confirms the model's ability to recover thin, peripheral vessels while maintaining a constant memory footprint, making it a feasible solution for high-resolution 3D medical imaging.

While the current model demonstrates robustness on the TopCoW dataset, future research must focus on validating generalizability across larger, multi-center datasets. Additionally, we aim to explore "smarter" attention-based fusion strategies to dynamically weigh anatomical versus topological signals, and investigate increasing the dynamic stream's capacity to resolve even more complex ambiguities.

\section*{Acknowledgment}
This research was supported by the Croatian Science Foundation under the project number IP-2024-05-9492 and the European Regional Development Fund under grant agreement PK.1.1.10.0007 (DATACROSS).

\bibliographystyle{IEEEtran}
\bibliography{refs}  

\end{document}